# Simple Methods for Scanner Drift Normalization Validated for Automatic Segmentation of Knee Magnetic Resonance Imaging – with data from the Osteoarthritis Initiative

**Erik B. Dam**

- Biomediq A/S, Copenhagen, Denmark
- Department of Computer Science, University of Copenhagen
- Data Science Lab, Faculty of Natural Sciences, University of Copenhagen

**Abstract**. Scanner drift is a well-known magnetic resonance imaging (MRI) artifact characterized by gradual signal degradation and scan intensity changes over time. In addition, hardware and software updates may imply abrupt changes in signal. The combined effects are particularly challenging for automatic image analysis methods used in longitudinal studies. The implication is increased measurement variation and a risk of bias in the estimations (e.g. in the volume change for a structure). We proposed two quite different approaches for scanner drift normalization and demonstrated the performance for segmentation of knee MRI using the fully automatic KneeIQ framework. The validation included a total of 1975 scans from both high-field and low-field MRI. The results demonstrated that the pre-processing method denoted Atlas Affine Normalization significantly removed scanner drift effects and ensured that the cartilage volume change quantifications became consistent with manual expert scores.

**Address all correspondence to:** Erik B Dam, erikdam@di.ku.dk

## 1   Introduction

Variation in signal intensity distribution over time is a well-known effect for magnetic resonance imaging scans, and is typically termed scanner drift. For automatic methods, including segmentation and biomarker quantification, scanner drift may impact the measurements.

Ideally, analysis methods could be developed to be invariant to scanner drift. A classical example is registration using the normalized mutual information (NMI) measure to steer the optimization[1]. NMI is invariant to bijective intensity transformations. Another example is use of the scale-invariant feature transformation[2] (SIFT) to generate features used in voxel classification methods for segmentation. However, frameworks that implicitly or explicitly are invariant to intensity scaling will by definition ignore aspects of the intensity information. Therefore, it seems sensible to normalize the intensities by a simple transformation that preserves most of the information captured by the intensity levels.

Ideally, an intensity transformation could be based on a physical model of the actual sources of the intensity drift. However, these sources are numerous, including the scanner (equipment drift, updates of software, replacement of spare parts), environment (temperature, pressure, humidity), operator (patient positioning), intra-patient (liquid status, movement), and inter-patient variation. To directly model the interplay between these sources of intensity variation is a daunting task – particularly in the light of the large inter-subject biological variation that may mask the more subtle drift effects.

Another appealing approach could be to include a calibration object in the scan. This has traditionally been used to normalize intensity and geometry in traditional x-rays. It would also allow normalization with respect to geometric drift. However, for MRI, the construction of a phantom calibration object is much more challenging since it may be needed to include multiple "tissues" designed for each combination of MRI sequence and anatomical target. Even if it may be feasible to design such phantoms[3], they would need to fit within the coil together with the anatomy in order to be used for drift correction.

The focus of this paper is normalization of intensity variation. We will limit ourselves to two processes we denote *drift* and *shift*. The drift is a consequence of wear and tear and results in gradual loss of signal over time. Contrarily, shift is the result of hardware/software/environment changes that affect the overall signal abruptly, often improving the signal to compensate for a period of drift. The challenge is to compensate for drift and shift due to the scanner without normalizing away intensity variation related to biological variation including pathology. Specifically, for the application of knee MRI segmentation used in this work, the articular cartilage should not be normalized to a uniform intensity within a scan or across a collection of scans. We consider the following potential approaches that are illustrated in Fig 1:

a) *Normalize individual scan*

   The simplest normalization where each scan is, for instance, normalized to have an intensity distribution with zero mean and standard deviation one.

   The challenge here is that knee MRI have large variation in the joint tissues included in the scan due to differences in knee size and anatomy, as well as large differences in muscle and fat tissue present. Therefore, such normalization will likely introduce large variation in small tissues such as the central cartilage structures.

b) *Normalize individual scan using reference template or atlas*

   A common approach is to normalize the scan against a single template or a multi-atlas and several methods for global intensity transformations have been published. These methods typically perform histogram matching to a template and transform the intensities to normalize the histograms[4–6] (e.g. using a polynomial intensity transformation[4]). For brain MRI, a more sophisticated patch-based method for focal intensity correction was demonstrated[7]. This used a library of image patches from scans with homogeneous intensity and matched focal patches to replace the patch center intensity with the intensity from the best matching library patch. This method potentially handles noise removal, intensity inhomogeneity, and scanner drift. However, there is a risk of normalizing away pathologies and the authors suggest limiting the use to normal subjects. These methods were validated for brain MRI, except one validated for prostate analys[5].

   Straightforward histogram-driven normalization is problematic for knee MRI, where the leg is partially outside the scan and tissue distributions vary greatly due to biological variation, as mentioned above.

c) *Normalize baseline and follow-up scans relatively to each other*

   Particularly in brain MRI, where focal atrophy is a key pathology indicator, registration-based analysis is very common. This approach is also applicable to scanner drift normalization. When only longitudinal change analysis is required, this is an appealing approach since the large biological variation between subjects is not a challenge when two scans from the subject are analyzed together. Therefore, intensity drift as well as inhomogeneity field difference can be robustly corrected[8].

d) *Normalize quantified biomarkers across all scans in completed study*

   Instead of directly addressing the scanner drift in a pre-processing step, the drift effects could also be normalized in post-processing using the entire collection of scans. The inclusion of an entire study may allow robust estimation of the subtle *drift* and abrupt *shift* effects as a function of time for each site.

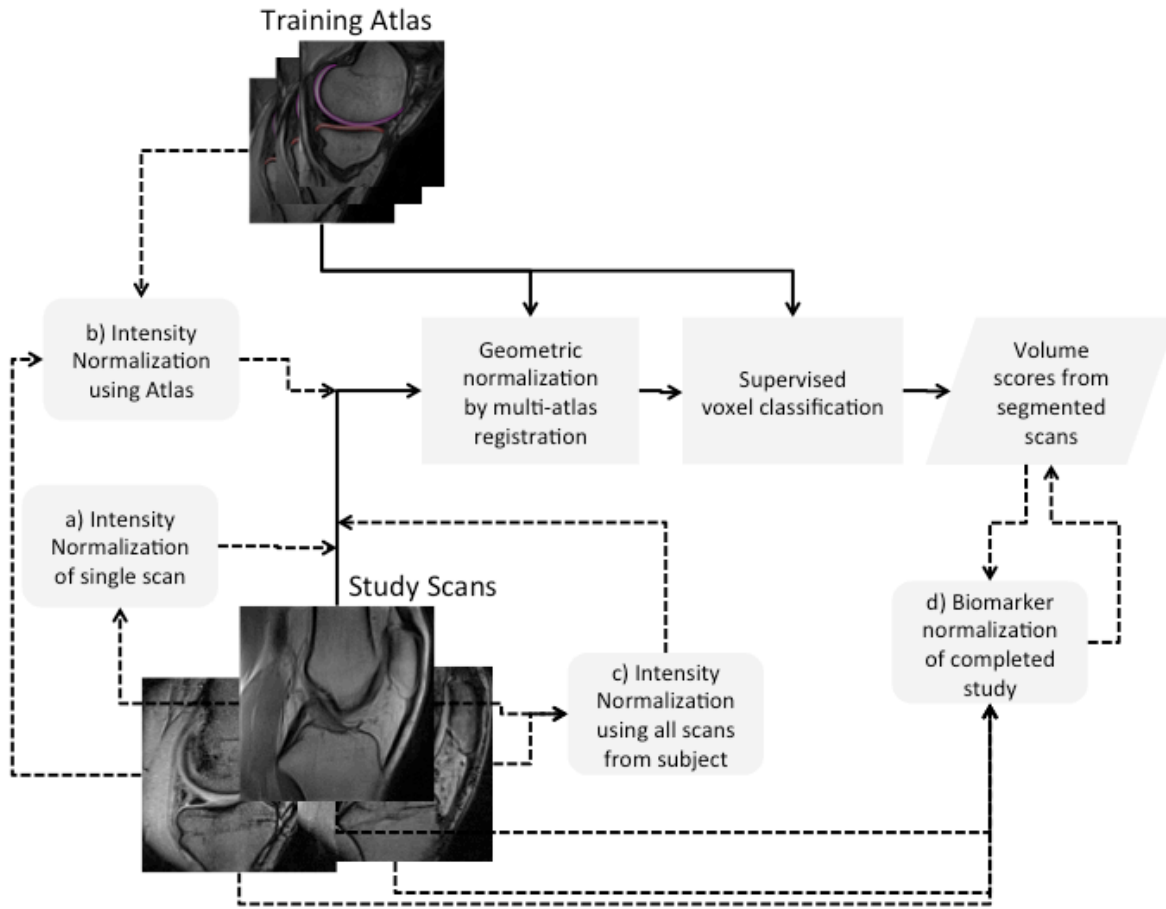

**Fig. 1**: Each of the scanner drift normalization options a)-d) are illustrated in the workflow of the KneeIQ framework that includes a multi-atlas registration followed by voxel classification to quantify cartilage volume scores for each scan in the study. The framework is trained using a collection of scans with manual expert annotations for the cartilage compartments. The solid lines show the KneeIQ data flow. The dotted lines show the pre-processing steps of normalization options a)-c) and the post-processing step of option d).

Among the four approaches above, we will not consider option c) since we aim to allow cross-sectional analysis of quantified biomarkers and not limit ourselves to longitudinal analysis. We previously implemented the trivial zero-mean, std-one normalization mentioned in option a) as an early experiment and experienced slightly worse segmentation accuracy compared to no normalization (these results are not reported). This experiment confirmed the challenge of tissue distribution variation in knee scans. As a result of the above, we propose two simple and very different methods for scanner drift normalization:

- A pre-processing method for normalization of each scan using a collection of training scans. To avoid the problem with tissue distribution variation, we apply a multi-atlas registration to ensure the intensities are analyzed robustly in a well-defined structural window. The method is termed *Atlas Affine Normalization* (AAN).
- A post-processing method that models the scan intensity drift and shift across the entire study using the scanner maintenance logs and normalizes the impact on the estimated biomarker scores (e.g. cartilage volumes) directly from this. This is termed *Piecewise Linear Drift Shift* (PLDS) correction.

We implemented these methods and demonstrated them in combination with a recently published, fully automatic method for segmentation of knee MRI[9]. The validation included two longitudinal studies of the progression of knee osteoarthritis (OA). Knee OA is the most dominant from of arthritis and among the highest contributors to disability in the world[10]. A central imaging biomarker for knee OA is to measure the cartilage degradation as the volume change over time. Typically, only small yearly cartilage loss rates less than 1% are observed[11] meaning that measurement accuracy and precision are essential. Given a typical re-measurement precision of around 5% for cartilage volume scores[12], disease effects risk drowning in measurement variation. Furthermore, if scanner drift introduces a measurement bias, it may be impossible to observe cartilage-protective treatments effects. Therefore, control of scanner drift is essential for the task of cartilage volume quantification.

## 2 Methods

The methods section includes descriptions of the cohorts, the existing segmentation framework, the two drift normalization methods, and finally the experiments. The descriptions of the cohorts and the segmentation framework are brief, for more detail we refer to the recent KneeIQ segmentation paper[9].

### 2.1 Knee MRI Collections

We used cohorts of knee MRI from the Center for Clinical and Basic Research (CCBR) and the Osteoarthritis Initiative (OAI) study. The collections included scans with manual segmentations that we used for training and validation. The relevant ethical review boards approved the studies.

The collections differed substantially in scanner type, sequence design, and disease severity as described below. The scanner and sequence specifications are given in Table 1.

**Table 1**: Study descriptions in terms of scanners, sequences, and available manual annotations.

|  | CCBR | OAI |
|---|---|---|
| Scanner | Esaote 0.18T C-span | Siemens 3T Trio |
| Sequence | Sagittal Turbo 3D T1<br>40° flip angle,<br>50 ms RT, 16 ms ET,<br>0.7 x 0.7 mm pixels,<br>0.8 mm slices,<br>scan time 10 min | Sagittal 3D DESS WE<br>25° flip angle,<br>16 ms RT, 4.7 ms ET,<br>0.36 x 0.36 mm pixels,<br>0.7 mm slices,<br>scan time 10 min |
| Compartments with manual segmentations | Tibia Bone<br>Medial Tibial Cartilage (MTC)<br>Medial Femoral Cartilage (MFC) | Medial Tibial Cartilage (MTC)<br>Lateral Tibial Cartilage (LTC)<br>Femoral Cartilage (FC)<br>Medial Femoral Cartilage (MFC)<br>Lateral Femoral Cartilage (LFC)<br>Patellar Cartilage (PC)<br>Medial Meniscus (MM)<br>Lateral Meniscus (LM) |

The CCBR cohort included 159 subjects in a longitudinal study of 21 month. Approximately half the subjects had no baseline radiographic signs of OA, defined as Kellgren and Lawrence[13] (KL) score 0, and the rest were evenly distributed from KL 1 to KL 3. From this cohort, 30 and 100 knee MRIs were used for training and validation, respectively. These MRI also had manual segmentations performed by an experienced radiologist. Among the validation knees, 38 also had rescans a week after baseline to evaluate scan-rescan variation.

The OAI cohort included 4796 subjects and six main, yearly visits, from which we included two sub-cohorts of 88 and 565 subjects and analyzed visits V00 and V01.

The small OAI sub-cohort mainly included KL 2 and KL 3 knees, similar to a typical phase 3 trial population. The large sub-cohort also had around one third knees that are KL 0 or KL 1. The 88 subjects were semi-manually segmentation by iMorphics and are available via the OAI website[14]. We divided the baseline scans for this set into 44 scans for training and 44 for validation. Further, we split the femoral cartilage segmentations into medial and lateral sub-compartments for the training set.

The large OAI sub-cohort was selected as OAI project 9B[14], where cartilage volume scores for the double-echo steady-state (DESS) scans made by Chondrometrics are publicly available[14]. This set includes 565 knees with volumes for both V00 and V01.

Both cohorts were evenly distributed among men and women and between left and right knees. For details on the cohorts, see the segmentation paper[9].

### 2.2 Existing KneeIQ Segmentation Framework

The segmentation framework used for the experiments combines rigid multi-atlas registration with voxel classification in a multi-structure setting[9]. The framework is trained from a collection of scans with expert segmentations of a set of anatomical structures. The framework name "KneeIQ" comes from Knee Image Quantification.

The multi-atlas registration step allows geometric transformation of each scan to a common space where structures are located consistently. This allows focused training of a voxel classifier for each structure. The classifier is based on features selected for each structure among a large set of potential features computed for each voxel including position, intensity, the 3-Jet at 3 scales[15], and non-linear combinations of derivatives such as the Hessian eigenvalues. This defines a k-nearest neighbors (kNN) classifier for each anatomical structure. The segmentation of the full scan is defined by choosing the structure with strongest classifier response for each voxel.

Since neither intensity nor the 3-Jet features are invariant to scan intensity changes, the framework may be sensitive to scanner drift. Particularly, an overall scanner drift between baseline and follow-up visits may create a general over- or under-segmentation bias for the follow-up scans.

#### 2.2.1 Multi-Atlas Rigid Registration

The aim of the registration step in KneeIQ is to produce a geometric transformation from a given scan to a common atlas space. Since we augmented this step to perform scanner drift normalization in one of the proposed methods, we describe the registration step briefly.

In the multi-atlas registration, a study scan is registered to each atlas scan using multi-scale rigid registration where the optimization criterion is NMI. To avoid boundary extrapolation artifacts, an outer margin of 15% of the scan is excluded in the evaluation of NMI. Each registration results in a transformation consisting of translation, rotation and scaling parameters. The final transformation for the scan is defined as the element-wise median. Since the registration is driven by NMI, this step is invariant to bijective intensity transformations.

This registration is summarized as follows. During training, each atlas scan $Im^t$ is registered to all other atlas scans and the median transformation $R^t$ gives the transform to the atlas center. When segmenting a new scan $S$, the scan is registered to all atlas scans resulting in transformations $S^t$. The compositions $S^t x R^t$ provide transformations to the atlas center via each atlas scan. Then element-wise median of these compositions defines the transformation from $S$ to the atlas center.

## 2.3  Scanner Drift Normalization: Affine Atlas Normalization

The Atlas Affine Normalization (AAN) augmented the multi-atlas registration step in KneeIQ. Given the rigid registration of a study scan to an atlas scan, the study scan was transformed using the geometric parameters.

We defined an affine intensity transformation as the least squares solution mapping the intensities from the transformed scan to the atlas scan – where the 15% periphery of the scans were ignored to avoid boundary extrapolation artifacts (illustrated in Fig 2). The affine intensity transformation was defined as the element-wise median of the transformations to each atlas scan.

As done for the geometric transformation outlined above, this normalization step was first performed for each atlas scan. This produced an affine intensity transformation for each atlas scan $A^t$, $scale_t \cdot A^t + offset_t$, and an affine transformation for the study scan $S$ to the atlas $A^t$, $scale_S \cdot S + offset_S$. The transformation for the study scan over the atlas scan to the atlas space center was the combination of the two transformations: $scale_t \cdot (scale_S \cdot S + offset_S) + offset_t$.

We defined the final affine intensity transformation for the study scan as the median of the combination transformation over each atlas scan. This affine transformation was applied to the study scan as a pre-processing step to the voxel classification step in the KneeIQ framework.

The AAN is not strictly a pre-processing stop to KneeIQ since it is applied between the two main operations of KneeIQ. However, since the multi-atlas registration is invariant to a global, affine intensity transformation due to the NMI criterion, we conceptually think of AAN as a pre-processing step.

As illustrated in Fig. 2, by ignoring a 15% margin, we excluded boundary artifacts as well as part of the muscle and fat tissues.

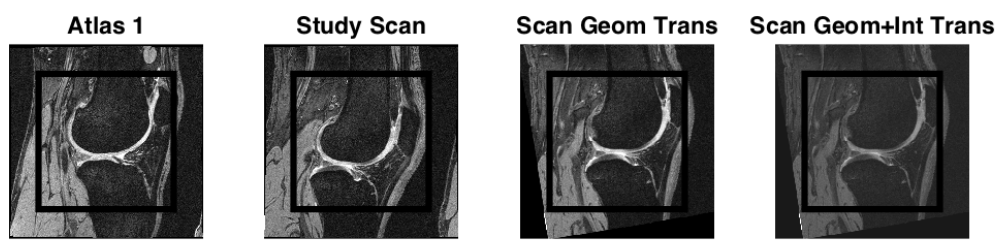

**Fig. 2**: The Atlas Affine Normalization. A Study Scan from the OAI V01 visit is transformed via the first atlas scan (Atlas 1). The geometric transformation is applied to the scan (Scan Geom Trans) and the least squares affine mapping of the scan intensities inside a 15% margin from scan to atlas is computed. This intensity mapping is combined with the mapping for the atlas scan to give the intensity mapping to the atlas space center. This final mapping is applied to the study scan (Scan Geom+Int Trans) as a preprocessing step before voxel classification.

*2.4  Scanner Drift Normalization: Piecewise Linear Drift-Shift*

It may not be possible to estimate scanner drift effects robustly from the individual scan due to the large biological variation between subjects. In this case, the drift may be modeled by assessing the overall change in intensities over time including all scans in the collection.

We designed the Piece-wise Linear Drift-Shift (PLDS) method under the assumption that scanner drift results in a slow, linear change in overall intensity over time, interrupted by abrupt shifts in overall intensity. Further, we assumed that the intensity changes are linearly related to changes in the measured imaging markers. Finally, the method assumed the availability of a scanner maintenance log to identify the dates where shifts can occur. Alternatively, header information in the study scans may be used to detect shift events such as scanner software updates.

The method is illustrated by the example in Fig. 3. The first step was to optimize a piece-wise linear model for each site of the scan mean intensity as a function of the scan day. We optimized using the Matlab "fmincon" function with the sum of absolute difference to the model as penalty function. This model described the population-averaged scanner drift and shift. The second step was to model the impact of intensity changes on biomarkers. This was done by modeling the linear relationship between scan mean intensity changes and biomarker changes.

The normalization using the PLDS method was then done for each scan by using the scan day and the biomarker values as the input. The scan day defined the change in intensity from study start using the piece-wise linear intensity model. The biomarker values were then corrected by the linear biomarker changes as function of the given intensity change.

For the CCBR study, the scanner maintenance log were available, defining dates for software updates and replacement of spare parts.

For the OAI scans, we used the attributes StationName and SoftwareVersion from the scan DICOM headers to search for shift events. For all four OAI sites, the software was updated at least once. For one OAI site, the station was updated. In addition to these automatically discovered shift events, we defined two speculative shift events by inspection of the intensity drift visualizations (see Fig. 3).

*2.5  Experiments and Statistical Analysis*

The AAN was used during pre-processing and affected the segmentations directly. Therefore it could be evaluated in terms of segmentation accuracy and precision. The PLDS was used as post-processing and could therefore only be validated on the resulting volume measurements.

We evaluated segmentation accuracy by the Dice volume overlap and as the linear correlation between biomarker scores, depending on availability of validation segmentations or just volume scores. Precision was evaluated by the root mean squared coefficient of variation (RMS CV) of the cartilage volume pairs from segmentations of repeated scans from the same subject acquired within a short period (scan-rescan precision). Such scan-rescan pairs were only available for the CCBR study.

The central validation of the drift normalization methods was to analyze the longitudinal changes for the biomarkers. For the OAI cohort, validation cartilage volumes were available, and the distributions of biomarker change scores were compared using a paired t-test. For the CCBR cohort, no validation volume scores were available at follow-up, and the volume changes were evaluated by comparing to scores typically reported in the literature.

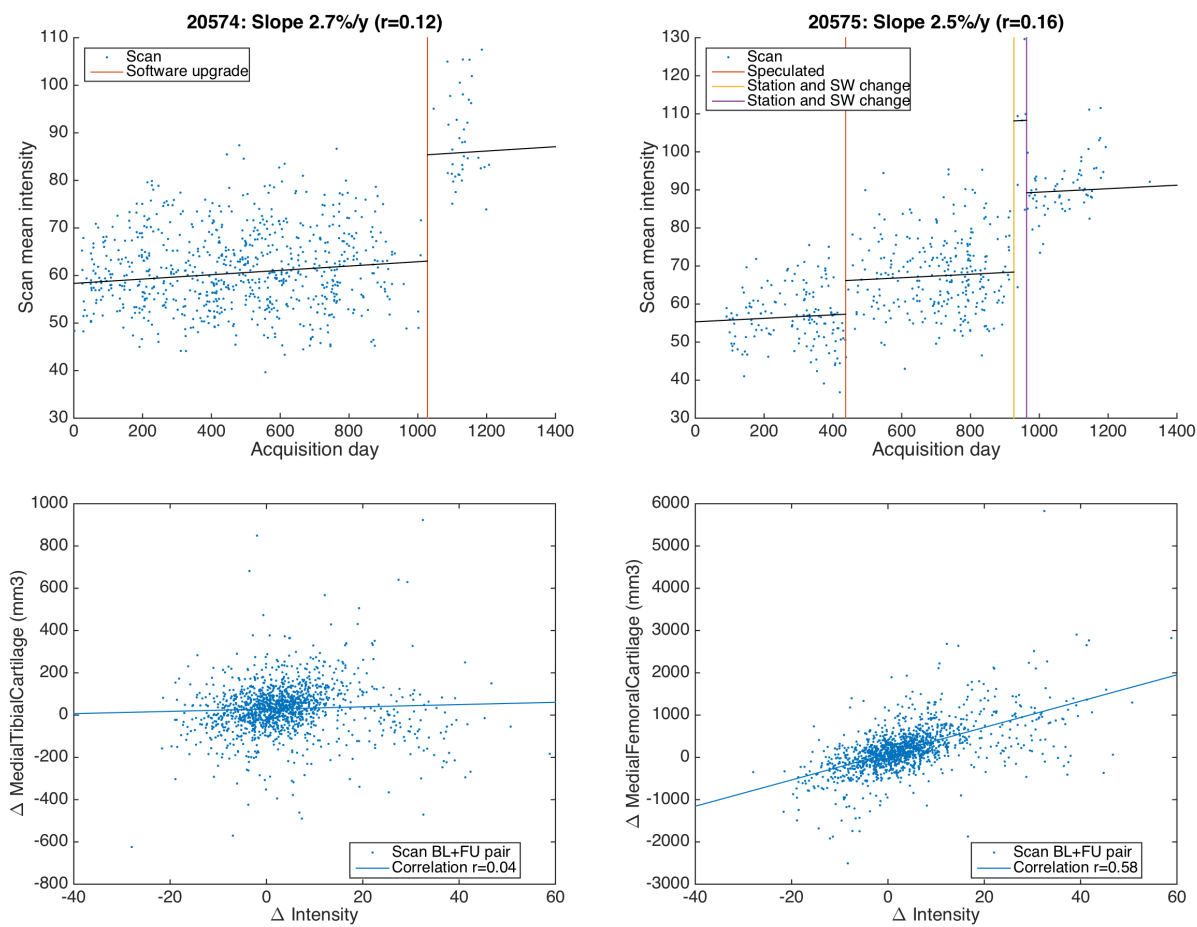

**Fig. 3**: The Piece-Wise Linear Drift Shift method. For each site, the relationship between scan day and scan mean intensity is modeled as piece-wise linear function. The shift events are marked by vertical lines. The top row shows the intensity models for two OAI sites. For each marker, the linear relationship between changes in intensity and changes in marker values are modeled. The bottom row shows this for two cartilage volume markers.

## 3 Results

The numbers of analyzed scans from each of the cohorts are listed in Table 2. The baseline (BL) scans were used for training and evaluation of segmentation accuracy and precision. The exception was the iMorphics set where the validation consisted of both BL and follow-up (FU) scans to investigate the effect of drift on the segmentation accuracy. For all collections, the pairs of BL and FU scans were used to quantify longitudinal changes and thereby investigate drift effects.

**Table 2**: Number of knee MRIs from each cohort used for training of the framework, for validation of segmentation accuracy and precision, and the number of BL/FU pairs for assessment of longitudinal changes. The iMorphics validation set included 44 knees at both V00 and V01 to evaluate drift in segmentation accuracy.

|  | Training | Validation | Rescans | Longitudinal |
|---|---|---|---|---|
| CCBR | 30 | 110 | 38 | 266 |
| OAI iMorphics | 44 | 44+44 |  | 44 |
| OAI Chondrometrics |  |  |  | 565 |

## 3.1 Segmentation Accuracy

The validation segmentations accuracies are shown in Table 3 for the CCBR and OAI iMorphics collections. The Dice volume overlaps were typically around 0.82, with patellar cartilage being the most challenging with scores around 0.75.

For comparison, for the CCBR study the radiologist repeated the manual segmentations for the 38 scans also used for the scan-rescan precision analysis. The intra-operator, intra-scan mean Dice volume overlaps for both medial cartilages were 0.86.

There was no clear difference in accuracy between the original KneeIQ framework and the version augmented with the Atlas Affine normalization. There was no clear different between accuracy for the V00 and V01 visits for the OAI sub-cohort.

**Table 3**: Segmentation accuracy for training and validation in the CCBR and OAI iMorphics collections. Accuracy is given as the Dice volume overlap showing mean±std values. The number of validation knees (n) is given for each set. The empty entries are for compartments with no available manual segmentations in each sub-cohort. The compartment abbreviations are defined in Table 1.

| Com-partment | CCBR (n=110) | | OAI V00 (n=44) | | OAI V01 (n=44) | |
|---|---|---|---|---|---|---|
| | KneeIQ | KneeIQ+AAN | KneeIQ | KneeIQ+AAN | KneeIQ | KneeIQ+AAN |
| MTC | 0.839±0.048 | 0.839±0.046 | 0.811±0.056 | 0.816±0.050 | 0.809±0.052 | 0.814±0.052 |
| LTC | | | 0.865±0.034 | 0.868±0.037 | 0.861±0.037 | 0.863±0.037 |
| MFC | 0.804±0.059 | 0.798±0.064 | 0.814±0.043 | 0.819±0.037 | | |
| LFC | | | 0.841±0.043 | 0.847±0.045 | | |
| FC | | | 0.838±0.037 | 0.843±0.036 | 0.836±0.032 | 0.839±0.030 |
| PC | | | 0.748±0.101 | 0.765±0.102 | 0.750±0.099 | 0.756±0.101 |
| MM | | | 0.758±0.085 | 0.754±0.081 | 0.754±0.088 | 0.751±0.089 |
| LM | | | 0.830±0.055 | 0.822±0.064 | 0.823±0.057 | 0.818±0.058 |
| Cartilages, Mean | 0.820±0.049 | 0.816±0.052 | 0.816±0.076 | 0.823±0.072 | 0.814±0.073 | 0.818±0.072 |

## 3.2 Cartilage Volume Precision

The scan-rescan precision for the CCBR collection is given in Table 4. For comparison, the radiologist had intra-scan precision 6-7% (repeated manual segmentation of same scan). For OAI, no scan-rescan examples are publicly available.

All variants of the automatic KneeIQ segmentation framework had better scan-rescan precision than the radiologist. The Atlas Affine drift normalization had better precision than the other variants for the medial tibial cartilage compartment, however this was not the case for the total tibial and femoral medial cartilage volume.

**Table 4**: Precision of cartilage volumes for the 38 scan-rescan knees from the CCBR collection. The scan-rescan results are for segmentations of repeated scans acquired approximately 1 week apart. Precision is given as the root mean squared coefficient of variation (RMS CV).

| Compartment | CCBR, Manual, Radiologist (n=38) | | | |
|---|---|---|---|---|
| | Radiologist | KneeIQ | KneeIQ+PLDS | KneeIQ+AAN |
| MTC | 8.9% | 4.9% | 4.9% | 3.8% |
| MFC | 9.2% | 4.9% | 4.8% | 5.1% |

| Total Cartilage | 6.1% | 4.1% | 4.1% | 4.1% |

### 3.3 Cartilage Volumes, Cross-sectional Agreements

The cross-sectional agreement between volumes for the baseline visit is given for each cohort in Table 5. The volumes from the variants of the automatic method were compared to the reference volumes from manual or semi-automated segmentations. There were slight variations in both linear correlation and bias offset across the automated KneeIQ framework and the two drift normalization methods, but no clear trends were apparent.

**Table 5**: Cross-sectional agreement between the automatic KneeIQ volume scores and the validation segmentation volumes on the CCBR collection and OAI sub-cohorts. The table includes number of knees available (n), Pearson linear correlation coefficient (r), and mean signed relative difference (%). Abbreviations are given in Table 1.

| Cohort | KneeIQ | | KneeIQ+PLDS | | KneeIQ+AAN | |
|---|---|---|---|---|---|---|
| Compartment | r | % | r | % | r | % |
| CCBR (Validation: Radiologist, manual, n=110) | | | | | | |
| MTC | 0.91 | -7 | 0.90 | -7 | 0.90 | -9 |
| MFC | 0.92 | -5 | 0.92 | -5 | 0.90 | -4 |
| Total Cartilage | 0.94 | -5 | 0.93 | -5 | 0.92 | -5 |
| OAI (Validation: iMorphics, semi-automatic, n=44) | | | | | | |
| MTC | 0.93 | 11 | 0.93 | 12 | 0.92 | 12 |
| LTC | 0.94 | 6 | 0.94 | 6 | 0.94 | 4 |
| MFC | 0.89 | 6 | 0.89 | 4 | 0.91 | 0 |
| LFC | 0.91 | 5 | 0.91 | 5 | 0.90 | 4 |
| FC | 0.92 | 5 | 0.92 | 4 | 0.93 | 2 |
| PC | 0.88 | 15 | 0.88 | 17 | 0.90 | 8 |
| MM | 0.85 | 12 | 0.85 | 14 | 0.87 | 11 |
| LM | 0.87 | 4 | 0.87 | 7 | 0.79 | 6 |
| Total Cartilage | 0.92 | 9 | 0.92 | 10 | 0.92 | 7 |
| OAI (Validation: Chrondrometrics, manual, n=565) | | | | | | |
| MTC | 0.92 | 11 | 0.92 | 11 | 0.92 | 14 |
| LTC | 0.87 | 17 | 0.87 | 17 | 0.88 | 17 |
| Total Cartilage | 0.90 | 14 | 0.90 | 14 | 0.90 | 15 |

### 3.4 Cartilage Volumes, Longitudinal Agreements

The longitudinal volume changes are listed in Table 6 for the three cohorts. The each cohort, the manual or semi-automatic validation segmentations defined a reasonable level for the mean volume change for each structure. For each variant of the automatic framework, the corresponding volume changes were compared to validation changes using a Student's t-test. Here, low p-values indicate that it was unlikely that the two methods captured the same distribution. Therefore, the automatic method and the drift normalizations should ideally result in high p-values, indicating that they captured the same volume changes as the validation scores.

For the CCBR cohort, no validation segmentations were available for the follow-up scans. These results were therefore qualitatively compared to the expected cartilage volume changes over 21 months, as suggested by previous studies. For the small OAI cohort, there was no clear pattern in agreement, neither across the structures nor across the method variants.

For the large OAI cohort, the validation segmentations indicated a moderate cartilage loss given by changes around -1%. The original KneeIQ method and the PLDS normalization both resulted in cartilage growths around 1%. The combination of KneeIQ and Atlas Affine normalization resulted in cartilage loss levels similar to the validation segmentations.

**Table 6**: Longitudinal cartilage volume changes for the automatic KneeIQ volume scores and the corresponding validation scores on the CCBR and the OAI sub-cohorts. The table includes number of knees available (n), mean relatively changes ± standard deviation (%). For each of the KneeIQ variants, the p-value from a t-test comparison to the validation scores is given, where p>0.05 indicates that the scores are not statistically different (large p value indicate the method agrees with the validation segmentations). The compartment abbreviations are in Table 1.

| Cohort | Validation | KneeIQ | | KneeIQ+PLDS | | KneeIQ+AAN | |
|---|---|---|---|---|---|---|---|
| Compartment | % | % | p | % | p | % | p |
| CCBR (Validation: Radiologist, manual, n=266) | | | | | | | |
| MTC | | 4.3% ± 10.0% | | 5.3% ± 10.1% | | 3.1% ± 8.9% | |
| MFC | | -0.2% ± 13.7% | | 1.5% ± 13.9% | | -1.8% ± 12.7% | |
| Total Cartilage | | 1.0% ± 11.6% | | 2.4% ± 11.7% | | -0.5% ± 10.4% | |
| OAI (Validation: iMorphics, semi-automatic, n=44) | | | | | | | |
| MTC | 0.2% ± 7.4% | 1.7% ± 6.4% | 0.2555 | 2.5% ± 6.3% | 0.0868 | 2.1% ± 6.3% | 0.1560 |
| LTC | -0.5% ± 5.4% | 1.2% ± 5.2% | 0.0955 | 0.2% ± 5.2% | 0.4634 | 0.8% ± 5.0% | 0.1221 |
| FC | 0.9% ± 4.2% | 1.8% ± 5.8% | 0.3922 | 0.4% ± 5.8% | 0.5671 | 2.0% ± 4.1% | 0.1732 |
| PC | -2.0% ± 8.3% | 0.2% ± 10.5% | 0.3131 | 4.2% ± 10.5% | 0.0070 | 2.6% ± 15.0% | 0.0932 |
| MM | 2.9% ± 10.0% | -3.0% ± 10.4% | 0.0143 | -0.4% ± 10.1% | 0.1540 | -2.6% ± 11.1% | 0.0199 |
| LM | 1.3% ± 6.8% | 1.3% ± 11.7% | 0.9630 | 6.2% ± 11.9% | 0.0045 | 0.3% ± 9.5% | 0.4533 |
| Total Cartilage | 0.3% ± 3.2% | 1.5% ± 4.9% | 0.1104 | 1.2% ± 4.9% | 0.2445 | 1.7% ± 3.8% | 0.0248 |
| OAI (Validation: Chrondrometrics, manual, n=565) | | | | | | | |
| MTC | -1.0% ± 4.9% | 1.0% ± 6.5% | <0.0001 | 1.3% ± 6.2% | <0.0001 | -0.4% ± 6.0% | 0.0152 |
| LTC | -1.4% ± 4.7% | 1.2% ± 6.7% | <0.0001 | 0.5% ± 6.6% | <0.0001 | -1.3% ± 6.2% | 0.7453 |
| Total Cartilage | -1.2% ± 3.5% | 1.0% ± 4.9% | <0.0001 | 0.8% ± 4.9% | <0.0001 | -0.9% ± 4.7% | 0.1999 |

## 4   Discussion

Two fundamentally different scanner drift and shift normalization methods were implemented and validated for the task of cartilage segmentation from knee MRI.

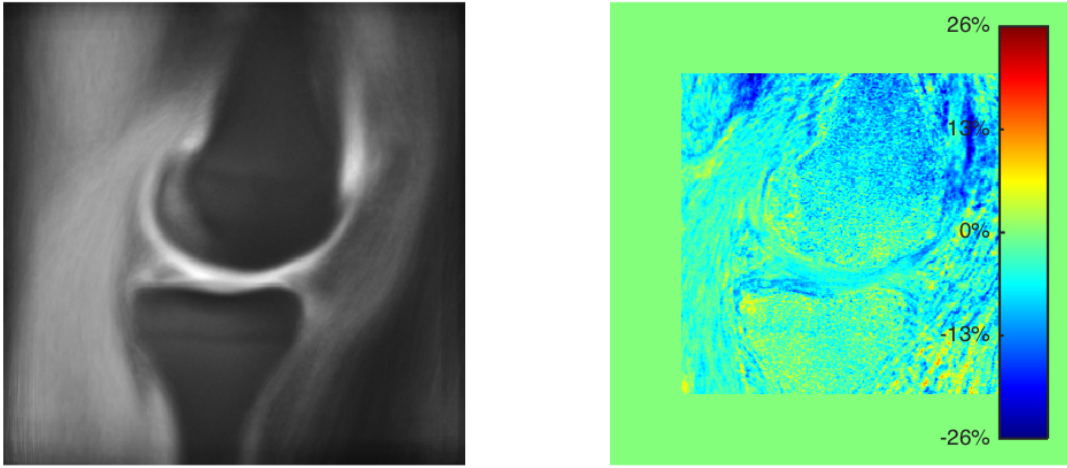

**Fig. 4**: Visualization of the intensity drift for an OAI site (station name 20576). Each BL scan was transformed to a common space using the registration transform and the mean intensity was computed at each position (left). The same was done for each FU scan, and the relative, focal intensity change was computed. To ignore boundary extrapolation effect, the 15% outer margin was cleared (right). For this site, there was a mean intensity loss of 3.3% from BL to FU, but the intensity changes were not uniformly distributed across the scan. Some (yellow) regions had intensity increase, whereas most (green-blue) regions had some loss.

## *4.1 Limitations*

The simple, affine, global normalization methods that we investigated have strong limitations. The first is the assumption that the drift is global for a scan. For PLDS, there was further the assumption that the drift impact on the measured biomarkers is linear. Whether or not these assumptions are sufficiently close approximations of the actual drift depends on the scanner, the sequence, and the biomarker quantification method. As a consequence, they may be appropriate for one study, but not for others.

Specifically, the drift may result in changes in noise rather than changes in intensity levels. Or changes in scanner inhomogeneity. These interplay between these factors was previously addressed for the task of analysis of prostate cancer from MRI[5]. Their results demonstrated that scanner drift normalization alone worsened the classification results, these results improved when performing combination of bias field correction and noise filtering prior to intensity standardization.

Further, the drift may have different effects on different tissue, again depending on the sequence. The illustration in Fig. 4 hints that this is to some degree the case for the OAI scans.

In addition, the observed changes in intensity may not be the consequence of scanner drift, but rather of changes in the subject – either due to pathology or changes in body composition. In these cases, drift normalization may introduce added measurement variation. Particularly, PLDS may respond to changes in body composition since added weight will "fill" the air around the knee in the scan field of view. Since AAN excluded a boundary margin from the drift estimate, the risk of weight dependency was smaller.

In general, due to the non-physical modeling of the scanner drift, our proposed normalization schemes will at best capture part of the scanner drift effects. The intention of the evaluation is to determine whether the methods improve the resulting biomarkers in terms of accuracy, precision, and particularly longitudinal change consistency. However, the evaluation is itself possibly slightly confounded by scanner drift effects since the validation segmentations may also be affected by scanner drift. Since manual segmentation is performed with the ability to change the intensity windowing interactively, the effect of simple changes in intensity levels should be minimal. However, the iMorphics segmentations that were performed using a semi-automated live-wire method may to some degree suffer from scanner drift. Due to this risk and the small size of the iMorphics sub-cohort, we have chosen to put less focus on the change measurement in this group. However, they are included in Table 6 for completeness.

*4.2  Visualizing the Scanner Drift*

The assumptions regarding the nature of the observed that be visually inspected. Since all scans were registered to the atlas space, we can compute the mean BL and the mean FU scans, and visualize the focal changes in intensity (see Fig. 4). The figure demonstrates that intensity loss occurred throughout the scan, but also that the distribution was not uniform.

This means that our drift normalization methods are likely to simplistic to fully capture these drift patterns since the methods assumed a global, affine drift. As a consequence, more sophisticated methods with spatially varying correction could in principle provide better drift normalization. However, the development of such methods is challenging since they could normalize away focal pathology. The evaluation demonstrated to what degree our simplistic assumptions approximated the observed drift and shift sufficiently to improve the cartilage volume quantifications.

*4.3  Results*

Both PLDS and AAN scanner drift normalizations could potentially improve consistency in segmentations when the collection is acquired over longer time periods. The AAN scanner drift may also improve robustness against fluctuations in signal occurring over the course of a day and thereby potentially improve scan-rescan precision. Here, we summarize the results for each method.

The original KneeIQ methods had good accuracy with Dice volume overlaps around 0.82 and volume linear correlations around 0.9 compared to validation segmentations. The scan-rescan precision was 4.9% for both evaluated compartments. However, the volume change measurements differed significantly from the manual validation segmentation scores, with mean tibial cartilage volume changes of +1.0% compared to -1.2% for KneeIQ and validation, respectively.

The PLDS normalization provided similar cross-sectional volume correlations and scan-rescan precision scores as the original KneeIQ. The volume change measurements were also significantly different from the manual validation scores.

The AAN normalization provided similar accuracy, but improved scan-rescan precision in the medial tibial compartment to 3.8%, albeit with a slightly worse precision in the femoral compartment at 5.1%. Regarding longitudinal volume changes, the AAN normalization resulted in scores similar to the manual validation set with mean total cartilage volume change of -0.9% compared to -1.2% from the manual segmentations.

*4.4  Segmentation of Meniscus*

The focus in this study was articular cartilage and robust estimation of volume changes. However, for the OAI study, we also quantified medial and lateral meniscus volumes. We have not discussed these results as the only longitudinal validation data was the small OAI sub-cohort with semi-automated segmentations that may also be confounded by scanner drift effects. Future research will show the validity of the meniscus segmentation and the meniscal volume changes.

### 4.5 Why AAN worked and PLDS did not?

Both AAN and PLDS relied on simple affine modeling of the drift effects. However, there were two fundamental differences. First, the intensity modeling for AAN was performed in a window where anatomical alignment was ensured by multi-atlas registration. This limited the impact of the differences in body composition and scan background that confounded the PLDS model. Secondly, the post-processing modeling by PLDS assumed a linear relationship between intensity changes and biomarker changes. This assumption may be too strong for many biomarkers.

### 4.6 Conclusions

There are complex, highly non-linear processes leading to scanner drift and subsequent changes in biomarkers quantified from the scans using automated analysis. In this study, we investigated whether two simple, linear methods could approximate the drift effects sufficiently to improve biomarkers in terms of accuracy, precision, and most importantly change measurements.

The post-processing PLDS method retained accuracy and precision, but was unable to improve the change measurements. The pre-processing AAN method retained accuracy, marginally improved precision, and improved longitudinal cartilage volume measurements. The resulting KneeIQ+AAN total cartilage volume changes of -0.5% and -0.9%, for the CCBR and OAI studies respectively, are in line with the literature and were very similar to the manual segmentation scores. Furthermore, the standard deviations on the change scores in Table 6 show that the measurement variation was also improved due to the AAN correction.

Thereby, the evaluation showed that the AAN method was apparently able to model scanner drift effects sufficiently well to improve the results significantly. And the improvement was sufficient to ensure that the resulting biomarker scores became very similar to manual expert scores in both cross-sectional and longitudinal comparisons. Future research will focus on the relationship between the volume changes and clinical observations with the hypothesis that the AAN scanner drift normalization will allow stronger clinical conclusions.

## 5 Acknowledgements

We gratefully acknowledge financial support from the Danish Research Foundation ("Den Danske Forskningsfond").

The research leading to these results has received funding from the D-BOARD consortium, a European Union Seventh Framework Programme (FP7/2007-2013) under grant agreement n° 305815.

The CCBR collection was provided by the Center for Clinical and Basic Research, Ballerup, Denmark, with manual cartilage segmentations by radiologist Paola C. Pettersen.

The OAI collection was provided by the Osteoarthritis Initiative with the cartilage and menisci segmentations performed by iMorphics. The OAI is a public-private partnership comprised of five contracts (N01-AR-2-2258; N01-AR-2-2259; N01-AR-2-2260; N01-AR-2-2261; N01-AR-2-2262) funded by the National Institutes of Health, a branch of the Department of Health and Human Services, and conducted by the OAI Study Investigators. Private funding partners include Merck Research Laboratories; Novartis Pharmaceuticals Corporation, GlaxoSmithKline; and Pfizer, Inc. Private sector funding for the OAI is managed by the Foundation for the National Institutes of Health. This manuscript was prepared using an OAI public use data set and does not necessarily reflect the opinions or views of the OAI investigators, the NIH, or the private funding partners.# 6    References

15Foundation for the National Institutes of Health. This manuscript was prepared using an OAI public use data set and does not necessarily reflect the opinions or views of the OAI investigators, the NIH, or the private funding partners.

# 6    References

1. D. Rueckert, A. F. Frangi, and J. A. Schnabel, "Automatic construction of 3-D statistical deformation models of the brain using nonrigid registration," IEEE Tranon Med. Imaging **22**(8), 1014–1025 (2003).
2. D. G. Lowe, "Object recognition from local scale-invariant features," in The Proceedings of the Seventh IEEE International Conference on Computer Vision, 1999 **2**, pp. 1150–1157 vol.2 (1999) [doi:10.1109/ICCV.1999.790410].
3. K. Yoshimura et al., "Development of a tissue-equivalent MRI phantom using carrageenan gel," Magn. Reson. Med. **50**(5), 1011–1017 (2003) [doi:10.1002/mrm.10619].
4. S. Prima et al., "Statistical Analysis of Longitudinal MRI Data: Applications for Detection of Disease Activity in MS," in Medical Image Computing and Computer-Assisted Intervention — MICCAI 2002, T. Dohi and R. Kikinis, Eds., pp. 363–371, Springer Berlin Heidelberg (2002).
5. D. Palumbo et al., "Interplay between bias field correction, intensity standardization, and noise filtering for T2-weighted MRI," Conf. Proc. Annu. Int. Conf. IEEE Eng. Med. Biol. Soc. IEEE Eng. Med. Biol. Soc. Annu. Conf. **2011**, 5080–5083 (2011) [doi:10.1109/IEMBS.2011.6091258].
6. L. G. Nyúl, J. K. Udupa, and X. Zhang, "New variants of a method of MRI scale standardization," IEEE Trans. Med. Imaging **19**(2), 143–150 (2000) [doi:10.1109/42.836373].
7. S. Roy et al., "Intensity inhomogeneity correction of magnetic resonance images using patches," 2011, 79621F – 79621F – 6 [doi:10.1117/12.877466].
8. B. Zou et al., "Simultaneous Registration and Bilateral Differential Bias Correction in Brain MRI," in Intelligent Imaging: Linking MR Acquisition and Processing (2015).
9. E. B. Dam et al., "Automatic segmentation of high- and low-field knee MRIs using knee image quantification with data from the osteoarthritis initiative," J. Med. Imaging **2**(2), 024001–024001 (2015) [doi:10.1117/1.JMI.2.2.024001].
10. M. Cross et al., "The global burden of hip and knee osteoarthritis: estimates from the Global Burden of Disease 2010 study," Ann. Rheum. Dis., annrheumdis – 2013–204763 (2014) [doi:10.1136/annrheumdis-2013-204763].
11. W. Wirth et al., "Comparison of 1-year vs 2-year change in regional cartilage thickness in osteoarthritis results from 346 participants from the Osteoarthritis Initiative," Osteoarthr. Cartil. OARS Osteoarthr. Res. Soc. **19**(1), 74–83 (2011) [doi:10.1016/j.joca.2010.10.022].
12. E. Schneider et al., "Equivalence and precision of knee cartilage morphometry between different segmentation teams, cartilage regions, and MR acquisitions," Osteoarthritis Cartilage **20**(8), 869–879 (2012) [doi:10.1016/j.joca.2012.04.005].
13. J. H. Kellgren and J. S. Lawrence, "Radiological assessment of osteo-arthrosis," Ann Rheum Dis **16**(4), 494–501 (1957).
14. "The Osteoarthritis Initiative (OAI)," Publicly available at www.oai.ucsf.edu, 2012, <www.oai.ucsf.edu>.
15. L. Florack, *The Syntactical Structure of Scalar Images*, PhD dissertation (1993).